\documentclass[runningheads,a4paper]{llncs}
\usepackage[utf8]{inputenc}
\usepackage{graphicx}
\usepackage{subcaption}
\usepackage{svg}
\usepackage{amsmath}
\usepackage{siunitx}
\usepackage{adjustbox}
\usepackage{hyperref}
\usepackage[noabbrev]{cleveref}
\usepackage{multirow,booktabs,makecell,array}
\usepackage{tikz,pgfplots,pgfplotstable}
\usetikzlibrary{patterns}
\pgfplotsset{compat=1.17}
\usepgfplotslibrary{groupplots,fillbetween,statistics}

\newcommand{\linewitherror}[5]{
    \addplot [each nth point=20,name path=minuserror,draw=none,no markers,forget plot] table [x={#2},y expr=\thisrow{#3}-\thisrow{#4}] {#1};
    \addplot [each nth point=20,name path=pluserror,draw=none,no markers,forget plot] table [x={#2},y expr=\thisrow{#3}+\thisrow{#4}] {#1};
    \addplot [forget plot,fill=#5,opacity=0.1] fill between[on layer={},of=pluserror and minuserror];
    \addplot [each nth point=20,#5,thick,no markers,legend image code/.code={\fill [fill=#5, opacity=0.1, draw=none] (0mm,-1ex) -- (0mm,1ex) -- (6mm,1ex) -- (6mm,-1ex) -- cycle; \draw [#5,thick] (0mm,0mm) -- (6mm,0mm);}] table [x={#2},y={#3}] {#1};
}

\def\addlegendimage{\csname pgfplots@addlegendimage\endcsname}

\definecolor{cola1}{RGB}{228,26,28}
\definecolor{cola2}{RGB}{55,126,184}
\definecolor{cola3}{RGB}{77,175,74}
\definecolor{cola4}{RGB}{152,78,163}
\definecolor{cola5}{RGB}{255,127,0}
\definecolor{cola6}{RGB}{255,255,51}
\definecolor{cola7}{RGB}{166,86,40}

\definecolor{colb1}{RGB}{102,194,165}
\definecolor{colb2}{RGB}{252,141,98}
\definecolor{colb3}{RGB}{141,160,203}
\definecolor{colb4}{RGB}{231,138,195}

\definecolor{colb1}{RGB}{102,194,165}
\definecolor{colb2}{RGB}{252,141,98}
\definecolor{colb3}{RGB}{141,160,203}
\definecolor{colb4}{RGB}{231,138,195}

\pgfplotsset{/pgfplots/colormap={mapCol1}{rgb255=(255,255,255) rgb255={228,26,28}}}
\pgfplotsset{/pgfplots/colormap={mapCol2}{rgb255=(255,255,255) rgb255={55,126,184}}}
\pgfplotsset{/pgfplots/colormap={mapCol3}{rgb255=(240,255,240) rgb255={77,175,74}}}
\pgfplotsset{/pgfplots/colormap={mapCol4}{rgb255=(240,78,240) rgb255={152,78,163}}}
\pgfplotsset{/pgfplots/colormap={mapCol5}{rgb255=(255,255,255) rgb255={255,127,0}}}
\pgfplotsset{/pgfplots/colormap={mapCol6}{rgb255=(255,255,255) rgb255={255,255,51}}}
\pgfplotsset{/pgfplots/colormap={mapCol7}{rgb255=(255,255,255) rgb255={166,86,40}}}

\usetikzlibrary{positioning}

\newcommand{\vt}[1]{\adjustbox{}{\tikz{#1}}}
\def\l{1.1mm}%
\def\s{0.3mm}%

\tikzset{
    gauss/.pic={
        \draw[yscale=0.5,yshift=-3,xscale=0.025,domain=0:10,red] plot function{exp(-(x-5)*(x-5)/2/2/2.0)/sqrt(pi)/2};
    },
    pics/rawmatrix/.style args={#1 x #2 #3}{
        code={
            \foreach \x in {1,...,#1}{
                \foreach \y in {1,...,#2}{
                    \fill[fill=#3] ({(\l+\s)*(\x-1)},{(\l+\s)*(\y-1)}) rectangle ({\l+(\l+\s)*(\x-1)},{\l+(\l+\s)*(\y-1)});
                }
            }
        }
    },
    pics/matrix/.style args={#1 x #2 #3 #4}{
        code={
            \pic[local bounding box=m] at (0,0) {rawmatrix={#1 x #2 #3}};
            \node[below=.5ex of m,minimum height=6mm] {#4};
        }
    },
    rawlongvector/.pic={
        \path (0,0) pic {rawmatrix={3 x 1 gray}};
        \draw[densely dotted,black] ({(\l+\s)*3},\l/2) -- ({(\l+\s)*6-\s},\l/2);
        \path ({(\l+\s)*6},0) pic {rawmatrix={3 x 1 gray}};
    },
    pics/longvector/.style args={#1}{
        code={
            \pic[local bounding box=v] at (0,0) {rawlongvector};
            \node[below=.5ex of v,minimum height=6mm] {#1};
        }
    },
    gmm1/.pic={
        \path (0,0.25mm) pic {gauss};
        \path (3mm,0) pic {rawmatrix={5 x 1 gray}};
    },
    pics/rawgmm/.style args={#1}{
        code={
            \foreach \i in {1,...,#1}{
                \path (0,{(\i-1)*2mm+0.25mm}) pic {gauss};
                \path (3mm,{(\i-1)*2mm}) pic {rawmatrix={5 x 1 gray}};
            }
        }
    },
    pics/gmm/.style args={#1 #2}{
        code={
            \pic[local bounding box=gmm] at (0,0) {rawgmm={#1}};
            \node[below=.5ex of gmm,minimum height=6mm] {#2};
        }
    },
    pics/genome/.style args={#1 #2}{
        code={
        \node[draw,rectangle, minimum size=#1cm] at (1*#1,1){5};
        \node[draw,rectangle, minimum size=#1cm] at (2*#1,1){2};
        \node[draw,rectangle, minimum size=#1cm] at (3*#1,1){0};
        \node[draw,rectangle, minimum size=#1cm] at (4*#1,1){8};
        \draw[densely dotted,black] (4*#1,1) -- (5*#1,1);
        \node[draw,rectangle, minimum size=#1cm] at (6*#1,1){9};
    }
    }
}

\newcommand\copyrighttext{%
  \footnotesize \textcopyright 2022 License: CC-BY-NC-ND 4.0 http://creativecommons.org/licenses/by-nc-nd/4.0/}

\newcommand\copyrightnotice{%
  \begin{tikzpicture}[remember picture, overlay]
  \node[anchor=south,xshift=9pt,yshift=5pt,remember picture] at (current page.south) {\fbox{\parbox{\textwidth}{\copyrighttext}}};
  \end{tikzpicture}%
}

\begin{document}

\title{Quality Diversity Evolutionary Learning of Decision Trees}

\titlerunning{Quality Diversity Evolutionary Learning of Decision Trees}

\ifdefined\ANONYMOUS
\author{\emph{Authors omitted for double-blind review}}
\institute{
\emph{Affiliations omitted for double-blind review}\\
\emph{}\\
}
\authorrunning{}
\else

\author{
Andrea Ferigo\inst{1}\orcidID{0000-0003-1795-011X} \and
Leonardo Lucio Custode\inst{2,3}\orcidID{0000-0002-1652-1690} \and
Giovanni Iacca\inst{3}\orcidID{0000-0001-9723-1830}
}

\authorrunning{A. Ferigo et al.}

\institute{University of Trento\\
Department of Information Engineering and Computer Science\\
Via Sommarive 9, 38123 Povo (TN), Italy\\
\email{\{andrea.ferigo,leonardo.custode,giovanni.iacca\}@unitn.it}}
\fi

\maketitle
\copyrightnotice
\begin{abstract}
Addressing the need for explainable Machine Learning has emerged as one of the most important research directions in modern Artificial Intelligence (AI). While the current dominant paradigm in the field is based on black-box models, typically in the form of (deep) neural networks, these models lack direct interpretability for human users, i.e., their outcomes (and, even more so, their inner working) are opaque and hard to understand. This is hindering the adoption of AI in safety-critical applications, where high interests are at stake. In these applications, explainable by design models, such as decision trees, may be more suitable, as they provide interpretability. Recent works have proposed the hybridization of decision trees and Reinforcement Learning, to combine the advantages of the two approaches. So far, however, these works have focused on the optimization of those hybrid models. Here, we apply MAP-Elites for diversifying hybrid models over a feature space that captures both the model complexity and its behavioral variability. We apply our method on two well-known control problems from the OpenAI Gym library, on which we discuss the ``illumination'' patterns projected by MAP-Elites, comparing its results against existing similar approaches.
\keywords{Quality diversity \and MAP-Elites \and Explainability \and Decision trees \and Reinforcement Learning \and Grammatical Evolution \and Entropy}
\end{abstract}


\section{Introduction}
\label{sec:intro}
As Artificial Intelligence (AI) has become more pervasive in real-world applications, major concerns have arisen regarding the need for explanations of its outcomes and, possibly, its inner working \cite{gerlings2020reviewing}. This need is especially relevant in safety-critical applications, such as (but not limited to) healthcare, control systems, or financial regulatory systems, where the opaqueness of modern AI, mostly based on Deep Learning (DL), may pose serious issues. As such, the field of eXplainable Artificial Intelligence (XAI) has produced considerable research efforts in the past two decades \cite{barredo_arrieta_explainable_2020,guidotti2018survey,adadi2018peeking}.

While there is currently a rather heated debate between those who believe that, also because of lack of explanations, DL is ``hitting a wall'' \cite{marcus2018deep,marcus2022deep}, and those who instead rightly highlight the many successes of modern DL---especially in Computer Vision and Natural Language Processing---it is however quite clear that, to some extent, and especially in some domains, explanations are necessary and stakeholders do really need them to fully trust AI models \cite{langer2021we}.

As an alternative to the dominant paradigm of black-box DL-based models, some researchers have recently advocated the use of white-box (also called glass-box) models, such as, for instance, decision trees and rule-based systems \cite{rudin_interpretable_2021}, noting that in some cases they can obtain similar or even better performance \cite{rudin_stop_2019,rudin_why_2019}. Moreover, while often in black-box models only a posteriori explanations are possible, white-box models are ``explainable by design'', in that their transparent structure makes it possible to directly interpret (and, possibly, understand) their inner working and thus their outcomes.

Given the importance of interpretability, interpretable Reinforcement Learning (RL) has thus been identified as one of the current grand challenges in AI \cite{rudin_interpretable_2021}. In fact, several modern applications of AI are modelled as RL problems. For example, deep RL has recently been used for the magnetic control of tokamak plasmas in a nuclear fusion plant \cite{degrave2022magnetic}, and for the definition of optimal taxation policies \cite{doi:10.1126/sciadv.abk2607}. These two are, clearly, high-risk domains where the decisions made by the AI can have serious consequences on people's lives and, hence, interpretable models may be more appropriate. However, the complexity of some real-world problems may be too high to be captured by simple white-box models. For this reason, a promising direction in current AI attempts to break the dichotomy between black-box and glass-box models, by proposing hybrid models \cite{meyer2019hybrid}, e.g., based on neuro-symbolic AI \cite{garcez2020neurosymbolic,susskind2021neuro,sarker2021neuro}. 

While seminal works on hybrid AI date back to the late '90s-early 2000s, see, e.g., the works by Sun et al. \cite{sun1997learning,sun2006connectionist}, or the studies on Learning Classifier Systems \cite{holland1999learning}, there is nowadays a resurgence of interest in those models. In this sense, it is worth noting that many recent successes of DL, such as DeepMind's MuZero~\cite{schrittwieser2020mastering} and its predecessors, are actually based on hybrid models.

Previous research has mainly focused on the optimization of hybrid models, i.e., the goal of those studies was to find their optimal configuration to improve performance. Some instances of hybrid models have been obtained by combining Q-learning \cite{watkins1992q} with decision trees induced by Genetic Programming, as in \cite{custode2022interpretable}, or by Grammatical Evolution (in the following ``GE''), as in \cite{custode2020evolutionary,custode_co-evolutionary_2021,custode2022pandemics}. Another recent work \cite{hallawa_evo-rl_2020} combined instead behavior trees with RL. However, when one wants to analyze this kind of models, other features---different from their performance---can be of interest. For instance, two important dimensions can be the \emph{model complexity}, i.e., a (static) measure of the model's structure, and its \emph{behavioral variability}, i.e., a (dynamic) measure of the model's capability of showing different behaviors during the execution of a given task. The first aspect can be relevant because, in general, simpler models can be easier to interpret \cite{rudin_stop_2019,rudin_why_2019}. The second aspect can be relevant because a higher behavioral variability may indicate a better adaptation and a higher robustness of the model \cite{haarnoja2018soft}.

In this paper, we apply for the first time a quality diversity (QD) algorithm, namely the Multi-dimensional Archive of Phenotypic Elites (in the following, MAP-Elites, or just ``ME'') \cite{mouret2015illuminating}, to ``disentangle'' the relation between a hybrid model's performance, its complexity, and its behavioral variability. QD is an emergent trend in Evolutionary Computation that posits that some specific tasks, especially those that are characterized by deceptive objectives, can be solved more efficiently by algorithms that explicitly look for a diversification of the solutions found during the evolutionary process, rather than an explicit optimization of a given objective function \cite{pugh2016quality}. Successful examples of QD algorithms are Novelty Search \cite{lehman2011novelty}, and indeed ME. The latter, in particular, has been designed with the goal of ``illuminating'' (w.r.t. the objective quality) a given feature space defined by some specific features of interest: this is the use of the algorithm that we make here. Originally devised for robotic applications \cite{cully2015robots}, ME has been successfully applied to various problems related to games \cite{khalifa2018talakat,fontaine2019mapping}, logistics and scheduling \cite{urquhart2018optimisation,urquhart2019illumination}, neuroevolution \cite{colas2020scaling}, and constrained optimization \cite{fioravanzo2021map}. Other works tried to use ME for interactive optimization \cite{urquhart2019increasing}, or to automatically derive rules to describe the relationships between features and objective quality \cite{urquhart2021automated}. In \cite{dolson2019exploring}, ME has been used for the first time to analyze programs evolved by means of Genetic Programming w.r.t. two aspects of program architecture, namely the scope count (to measure program modularity) and the instruction entropy (to measure instruction diversity).

Here, we use ME to obtain a diverse collection of interpretable hybrid models composed of a decision tree combined with Q-learning on the leaves. These models are similar to the ones used in previous works \cite{custode2020evolutionary,custode2022interpretable,custode_co-evolutionary_2021,custode2022pandemics} that, however, focus on the model optimization and analyze, a posteriori, the model complexity. In the present work, instead, we explicitly define as features for ME: 1) a measure of behavioral variability (i.e., the entropy of the actions taken by the model during the episode) and 2) a measure of model complexity (i.e., the depth of the decision tree). 
To the best of our knowledge, the only works that addressed the quest for diversity in RL tasks are the recent papers \cite{zahavy2022discovering} and \cite{tjanaka2022approximating}. In \cite{zahavy2022discovering}, in particular, authors state that finding diverse solutions to a same RL problem can improve exploration, transfer, hierarchy, and robustness of agents. In that work, diversity is explicitly measured as the distance between the state occupancies of the policies in the obtained policy set, with agents controlled by actor-critic neural networks. Here, instead, we implicitly measure diversity in the aforementioned two-feature space, and we focus on decision trees rather than neural networks. In \cite{tjanaka2022approximating}, authors also identify the policy diversity as the key for robust RL agents. Similarly to our work, they also use ME; however, differently from us, they use gradient approximations and, most importantly, they project the policies onto a feature space made of domain-specific features (in the case of a simulated locomotion task). In our case, instead, we consider domain-agnostic features, and as such our method can be of more general applicability. Despite these differences, our work shares the same motivation of these two previous works, i.e., we consider the search for diverse policies as a way to reach higher robustness. On top of that, we add the important consideration concerning the interpretability of such models. In this regard, it is however important to rule out a possible misconception: neither diversity nor interpretability are, \emph{per se} objectives of the search (after all, we do not know \emph{a priori} if these aspects are in conflict or not with the model performance). On the contrary, we consider them as \emph{features}, hence the need for disentangling their relation with respect to the performance, and the use of ME.

We apply the proposed method on two well-known classic control problems from the OpenAI Gym library \cite{brockman2016openai}, namely Cart Pole and Mountain Car, and compare the results of ME with those obtained by GE. We purposely exclude from the comparison black-box models based on deep neural networks: in fact, our goal is not to compare the results of ME and GE with other state-of-the-art models (also because such comparison has already been performed in \cite{custode2020evolutionary}), but rather focus only on interpretable models and demonstrate the effectiveness of ME at generating more diverse models than GE. Overall, we show that, by leveraging the exploration capability of ME, we are able to ``illuminate'' the relationship between model performance, complexity, and behavioral variability much more effectively than GE.

The rest of the paper is structured as follows. The next section describes the proposed method. The numerical results are presented in~\Cref{sec:results}. Finally,~\Cref{sec:conclusions} concludes this work and suggests possible future works.


\section{Method}
\label{sec:method}

As introduced before, we aim to evolve decision trees using a combination of an evolutionary algorithm (EA) and RL. While the EA evolves the structure of the tree, the RL algorithm optimizes the actions taken by the leaves, as in \cite{custode2020evolutionary}.

In the following, we describe the individual encoding, the EAs used to evolve the trees, the RL technique that optimizes the action of the leaves, and how we evaluate the decision trees, describing also the tasks performed.

\subsection{Individual Encoding}
\begin{figure}[ht!]
 \begin{subfigure}[b]{0.40\textwidth}
 \adjustbox{max width=\textwidth}{
 \centering
 \begin{tabular}{cc}
 \toprule
 Rule & Production\\
 \midrule
 Root & \texttt{if} \\
 If & \texttt{if} \texttt{Condition} then \texttt{action} else \texttt{action} \\
 Condition & $\sum_{i=0}^{n_inputs} \texttt{const} \cdot \texttt{input}_i <$ \texttt{const}\\
 Action & \texttt{leaf} or \texttt{if} \\
 const & $[1,10]$, with step of $1$\\
 \bottomrule 
 \end{tabular}
 }
 \caption{Schema of the simplified grammar used in the example.}
 \end{subfigure}
 \hfil
 \begin{subfigure}[b]{0.55\textwidth}
 \def\hs{5mm}
 \centering
 \begin{tikzpicture}
 \node[ shape=rectangle, rounded corners,
 draw, align=center] (r){$6x < 3$}
 child {node [shape=rectangle, rounded corners, draw, align=center, xshift=-5pt] {Leaf 2} edge from parent [black] node (flabel) [left, xshift=-5pt] {False}}
 child {node [shape=rectangle, rounded corners, draw, align=center, xshift=5pt] {Leaf 1} edge from parent [black] node [right, xshift=5pt] {True}};
 \node[left=\hs of flabel] (g){\vt{\pic{genome={0.5 0}};}};
 \node[below= 1mm of g] {$\mathbf{g}$};
 \draw[->] (g) -- (flabel);
 \end{tikzpicture}
 \caption{How a given genotype $\mathbf{g}$ (left) is translated into a decision tree (right).}
 \end{subfigure}
 \caption{Illustration of the individual encoding. (a) A simplified grammar; (b) example of translation of a genotype $\mathbf{g}$ into a decision tree, using the grammar shown in (b), with $maxValue=10$, and, for simplicity, a single input. The translation works as follows: the first rule \emph{root} always produces an \emph{if} node, which is composed of a condition and two actions. The condition rule requires $2$ \emph{const} nodes. Each \emph{const} rule selects an integer value between $1$ and $10$, hence in this case $l = 10$. In the example shown in (a), the first two values of $\mathbf{g}$ are $5$ and $2$, which correspond, respectively, to the fifth and second element, i.e., $6$ and $3$. The next $2$ values of $\mathbf{g}$, used for the action rule, are $0$ and $8$. The action rule can produce a \emph{leaf} or an \emph{action}, hence in this case $l = 2$. Therefore, the calculation performed to select the production is $i = 0\mod2$ and $j = 8\mod2$: in both cases, the first element of the \emph{action} production rule (a \emph{leaf}) is produced. As both nodes are \emph{leaf} nodes, no other nodes are produced and the rest of the genotype is not used. With this procedure, the genotype is translated into the decision tree. Note that the final action performed by each leaf is not encoded in the genotype, but is optimized using RL during the task.}
\label{fig:ind_encoding}
\end{figure}
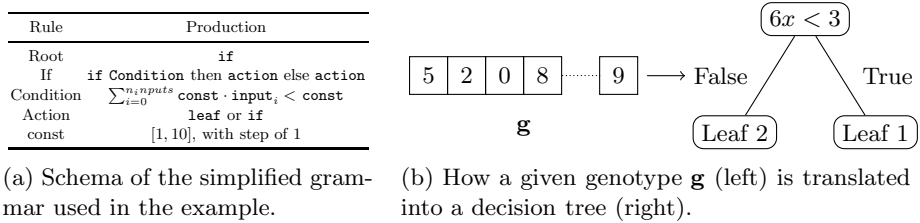

While the two algorithms (namely, GE and ME, as described below) used in this study are different, in both cases we encode the genotype of an individual (i.e., a candidate solution representing a decision tree) as a vector $\mathbf{g}= (g_0,\dots, g_{size})$ with $g_i \in [0, maxValue]$, where $maxValue$ is an integer value which must be greater than the number of possible choices for each production rule. We obtain the relative tree translating the genotype in the phenotype using an associate grammar \cite{goos_grammatical_1998}. This translation procedure operates as follows: given $l$ as the number of possible choices for a given production rule in the grammar, the value $c = \mathbf{g}_{i}\mod l$ indicates that the $c$-th value will be taken as value. In~\Cref{fig:ind_encoding} we show an example of such mapping with a simplified grammar. Note that we consider only oblique trees, i.e., trees in which each condition tests a linear combination of all the input variables.

\subsection{Evolutionary Algorithms}
The first EA we consider is a simple form of GE \cite{goos_grammatical_1998}. The second is a QD algorithm, ME \cite{mouret2015illuminating}, that aims to find an archive of different solutions rather than producing a single optimal solution.

For the two algorithms, we use the same mutation operator and the same computational budget, to make a fair comparison.

\subsubsection{Grammatical Evolution}
Following the basic form of GE \cite{goos_grammatical_1998}, we initially create a population of $n_{pop}$ randomly initialized solutions, then until we evaluate a total of $total_{pop}$ solutions, we repetitively create $n_{pop}$ new solutions. The new solutions are created as follows:
\begin{enumerate}
 \item we select $n_{pop}$ parents using tournament selection with size $k$;
 \item we group the solutions in pairs, and, with a probability of $p_{cx}$, we apply a crossover operator to each pair generating $2$ offspring that substitute the parents in the selection process;
 \item finally, each of the $n_{pop}$ solutions has a probability $p_{mu}$ to be mutated.
\end{enumerate}
Then, the new solutions are evaluated, and the best $n_{pop}$ solutions between the previous population and the new offspring are stored as the population for the next generation.
When all the $total_{pop}$ solutions are evaluated, the algorithm returns the best solution in the final population.

\subsubsection{MAP-Elites}
Multi-dimensional Archive of Phenotypic Elites~\cite{mouret2015illuminating}, commonly known as MAP-Elites, is a QD algorithm that maintains an archive of the best solutions that differ w.r.t. a given feature descriptor. 
The descriptor is necessary for ME to numerically describe each solution and, hence, store in the archive different solutions. Note that, while the descriptor needs to characterize a solution considering the problem being faced, it should be orthogonal to the solution's fitness (otherwise, if the selected features are highly correlated to fitness, the illumination pattern would be of little interest).

A descriptor is generally defined as a vector $\mathbf{d} = (d_0, \dots, d_n)$, with $d_i \in [min_i, max_i]$ and $\mathcal{D}: S \to \mathcal{R}^n$ being the function that, given a solution, returns its descriptor. 

The archive is an $n$-dimensional grid, with each dimension divided in $m$ bins. Thus, to find the coordinates $\mathbf{c}=(c_0,c_1,\dots,c_n)$ of a solution $s$ in the archive, we divide each dimension $i$ of the descriptor in $m$ equally wide bins, then we take the index of the bin in which the $d_i$ values fall as the $c_i$ coordinate.

During the evolution, to add a new solution $s$ to the archive, we calculate the coordinates $\mathbf{c_s}$ and, if the position in the map is empty, we insert the solution into the archive. Otherwise, if that position already contains a solution, we maintain only the one with the best performance. 

As each dimension is divided into $m$ bins, at the end of the evolution process the archive can store at most $m^n$ solutions. We populate the map in two steps: the initialization and the iterative phases. 
In the initialization phase, we randomly create $init_{pop}$ solutions and try to insert them into the archive. 
During the iterative phase, we repeatedly generate $batch_n$ solutions and try to add them to the archive, until we generate a total of $total_{pop}$ solutions (also including $init_{pop}$ solutions).
We create the new solutions as follows: 
\begin{enumerate}
 \item we randomly select $batch_n$ solutions from the archive;
 \item we mutate the $batch_n$ solutions.
\end{enumerate}

\subsubsection{Mutation}
We perform a uniform random mutation of each gene of an individual as follows: given the genotype ($\mathbf{g}$) of an individual, each gene has the same probability to be replaced with a new random gene with $g_{new} \in [0, maxValue]$. 

\subsubsection{Crossover} 
To perform crossover between a pair of parents $\mathbf{p}_1$ and $\mathbf{p}_2$, we proceed as follows.
We randomly select a point $i \in [0, size]$ in the genotype, split the parent vectors into two parts around the selected point, and then mix the parts into two new genotypes.

\subsubsection{Descriptor}
\label{subsec:descriptor}
As introduced before, ME stores a solution using a descriptor that indicates its position in the grid. In this work, we use two features to describe a decision tree. The first is a behavioral characterization of the decision tree, while the second represents the tree by its complexity.

To characterize the behavior of a solution, we use the entropy $E$ of the actions taken by the agent, calculated as follows: be $n\_actions$ the number of possible actions that the individual can perform and be $\mathbf{actions} = (a_0, \dots, a_{n\_actions})$ the vector of possible actions.
During the fitness evaluation, we store in $actions_{i}$ how many times the $i$-th action is performed by the tree.
Then, we calculate the vector of relative frequencies $\mathbf{f} = (f_0, \dots, f_{n\_actions})$ such that $f_{i} = \frac{actions_{i}}{\sum_{j=0}^{n\_actions} actions_{j}}$, from which we calculate the entropy $E = -\sum_{i=0}^{n\_actions}(f_{i} \cdot log_{n\_actions}f_{i})$. 
Note that, using as base for the logarithm the value $n\_actions$, $E$ takes values in $[0,1]$.
In this way a policy that makes always the same action will have an entropy of $0$, while a random policy, where $f_{i} = \frac{1}{n\_actions}$ for $i \in [0, n\_actions]$, will have an entropy of $1$. 

The second feature in the descriptor analyzes the structure of the tree, to characterize the solutions w.r.t. their complexity and, hence, their interpretability. For this reason, we use the depth of the tree, calculated after a simplification procedure carried out as in \cite{custode2020evolutionary}. This procedure simply consists in removing all the nodes that are not visited during the fitness evaluation. In this way, we produce a smaller tree pruned of all the nodes (including leaves) that are not used. Then, we calculate the depth of the simplified version of the tree. Note that simplifying the tree does not influence the values of the behavioral feature, as the actions that are pruned do not contribute to the entropy calculation since their frequency is null.

As mentioned in the introduction, we should stress once again that the two features defined above are not, \emph{per se}, objectives. As for entropy, it is not possible to state \emph{a priori} if this quantity should be minimized or maximized. In fact, it may well be that in some specific tasks a higher behavioral variability is to be preferred, while in others a less variable behavior may be better. For this reason, we consider entropy as a feature rather than an explicit objective. Likewise, one may in principle aim to explicitly minimize the tree depth, , to favor simpler models. However, it is difficult to state \emph{a priori} if the model complexity and its performance are conflicting goals or not: once again, it could be that in some cases simpler models actually perform better. For these reasons, using a multi-objective approach on these two quantities may be misleading, at best, or not appropriate at all. On the contrary, modeling them as features, and analyze a posteriori, through the illumination capability of MAP-Elites, their correlation with performance, appears to be a more suitable approach.

\subsection{Reinforcement Learning}
To optimize the action of the leaves, we use RL in the form of $\epsilon$-greedy Q-Learning \cite{watkins1992q}, with a fixed learning rate and a uniform random initialization.


\subsection{Fitness evaluation}
The evolved decision trees are used to solve control tasks. We independently test two OpenAI Gym \cite{brockman2016openai} environments, namely Cart Pole and Mountain Car.

For both environments, the procedure used to evaluate the decision tree is the following: the genotype is translated into the corresponding decision tree, then it is evaluated on $m$ independent episodes, where each episode uses a different seed for the random number generator.

Each episode is simulated until the task is solved or the time limit is reached. 
At each timestep, the reward from the environment is used to update the reinforcement model and it is accumulated for each episode. 
Moreover, in the Mountain Car environment, we normalize the observations in the range $[0,1]$ using the following formula: $\hat{x_i} = \frac{x_i - min_i}{max_i - min_i}$. The bounds used for the normalization are $[-1.2, 0.6]$ and $[-0.07, 0.07]$. We have found indeed that normalization is needed to solve this environment, while the Cart Pole task can be solved without.

When all the episodes have been simulated, the fitness of the individual is calculated as the average cumulative reward.

\subsubsection{Cart Pole}
In the Cart Pole environment\footnote{\url{https://gym.openai.com/envs/CartPole-v1/}} the agent has to maintain in equilibrium a pole over a cart.
At each timestep, the agent takes as input $4$ pieces of information: the position of the cart $x_c$, the velocity of the cart $v_c$, the pole angle $\theta_p$, and the pole angular velocity $\omega_p$. 
The agent can take $2$ actions: push the cart to the left or to the right. 
The reward is $+1$ for each timestep; each episode terminates after $500$ timesteps, or if $|\theta_p| > \ang{12}$ or if $|x_c| > 2.4$.
This task is solved if the cumulative reward for the agent has an average (on $100$ episodes) greater than or equal to $475$.

\subsubsection{Mountain Car}
\label{subsec:mc}
In the Mountain Car environment\footnote{\url{https://gym.openai.com/envs/MountainCar-v0/}} the agent has to move a car up a hill building up momentum thanks to another hill positioned before the car. 
The information available to the agent at each timestep is: the position along the x-axis of the car ($x_c$), and its velocity ($v_c$). 
At each step, the agent has $3$ possible actions: accelerate to the left, accelerate to the right, or do not accelerate. 
The task ends when the car reaches the top of the hill, or after $200$ timesteps. Until the car does not reach the top of the hill, the reward is $-1$ for each timestep; the task is considered solved if the average reward on $100$ episodes is greater than $-110$. 







\section{Results}
\label{sec:results}
In this section we present the results obtained by GE and ME on the two different tasks. We are foremost interested in comparing two aspects of the algorithms: the performance and the ``illumination'' capability.

For both algorithms, we performed $5$ independent runs to statistically verify the results. 
In~\Cref{tab:parametersGE} and~\Cref{tab:parametersME} we indicate the parameters used in the two environments with GE and ME respectively. Note that on the two tasks we use two different bounds for the entropy. In Mountain car, we set the bounds in the range $[0, 1]$, since three actions are possible. In Cart Pole, we instead set them in the range $[0.8, 1]$, as there are only two possible actions, and equilibrium between them is required to solve the task, i.e., solutions with lower entropy are quickly discarded.~\Cref{tab:grammar} describes the oblique grammar, which is common to all tasks and EAs. Finally,~\Cref{tab:rlParams} shows the parameters used by Q-learning.

As regards the interpretability of the solutions, previous works~\cite{custode2022interpretable,custode2020evolutionary,custode_co-evolutionary_2021} evaluate the complexity of the solutions based on the following factors: the number of symbols, the number of operations, the number of non-arithmetical operations, and how many times the non-arithmetical operations are consecutively composed. However, since in this work we use oblique trees, the complexity of each node is the same (as they all evaluate a linear combination of inputs, see the Condition rule in~\Cref{tab:grammar}). Hence, since total complexity of our evolved decision trees depends only on their depth, we use the latter as measure of complexity.

As for the ``illumination'' capability here we limit our analysis on a qualitative observation of how the two algorithms fill the feature space.

\begin{table}[ht!]
 \centering
 \begin{tabular}{
 lrr 
 }
 \toprule
 {Parameter} & {Cart Pole} & {Mountain Car} \\
 \midrule
 $n_{pop}$ & 200 & 200 \\
 $total_{pop}$ & 10000 & 200000 \\
 Tournament size & 2 & 2 \\
 $p_{cx}$ & 0.1 & 0.1 \\
 $p_{mu}$ & 1.0 & 1.0 \\
 Genotype size & 100 & 100 \\
 Genotype max value & 40000 & 40000 \\
 \bottomrule
 \end{tabular}
 \caption{Parameters used for GE.}
 \label{tab:parametersGE}
 \vspace{-1.2cm}
\end{table}

\begin{table}[ht!]
 \centering
 \begin{tabular}{
 lrr 
 }
 \toprule
 {Parameter} & {Cart Pole} & {Mountain Car} \\
 \midrule
 Bins for dimension & 10 & 10 \\ 
 Behavioral bounds & {$[0.8, 1.0]$} & {$[0, 1]$} \\ 
 Structural bounds & {$[1, 10]$} & {$[1, 10]$}\\
 $total_{pop}$ & 10000 & 200000 \\
 $batch_n$ & 20 & 20 \\
 $init_{pop}$ & 200 & 200 \\
 Tournament size & 2 & 2 \\
 $p_{cx}$ & 0 & 0 \\
 $p_{mu}$ & 1.0 & 1.0 \\
 Genotype size & 100 & 100 \\
 Genotype max value & 40000 & 40000 \\
 \bottomrule 
 \end{tabular}
 \caption{Parameters used for ME.}
 \label{tab:parametersME}
\end{table}

\begin{table}[ht!]
 \centering
 \begin{tabular}{cc}
 \toprule
 Rule & Production\\
 \midrule
 Root & \texttt{if} \\
 If & \texttt{if} \texttt{Condition} then \texttt{action} else \texttt{action} \\
 Condition & $\sum_{i=0}^{n_inputs} \texttt{const} \cdot \texttt{input}_i <$ \texttt{const}\\
 Action & \texttt{leaf} or \texttt{if} \\
 const & $[-1,1]$, with step of $0.001$\\
 \bottomrule 
 \end{tabular}
 \caption{Oblique grammar used in both EAs.}
 \label{tab:grammar}
\end{table}

\begin{table}[ht!]
 \centering
 \begin{tabular}{
 lrr
 S[table-format=2.3]
 S[table-format=2.1]
 }
 \toprule
 {Parameter} & {Cart Pole} & {Mountain Car} \\
 \midrule
 $\epsilon$ & 0.05 & 0.01 \\
 Initialization & Uniform $\in [-1,1]$ & Uniform $\in [-1,1]$ \\
 Learning Rate & 0.001 & 0.001\\
 Number of episodes & 100 & 100\\
 \bottomrule
 \end{tabular}
 \caption{Parameters used for the RL algorithm ($\epsilon$-greedy Q-learning).}
 \label{tab:rlParams}
\end{table}

\subsection{Cart Pole}
\begin{figure}[ht!]
 \centering
 \begin{tikzpicture}
 \begin{groupplot}[
 width=0.6\linewidth,
 height=0.4\linewidth,
 grid=both,
 grid style={line width=.1pt, draw=gray!10},
 major grid style={line width=.2pt,draw=gray!50},
 minor tick num=5,
 group style={
 group size=1 by 1,
 horizontal sep=4mm,
 vertical sep=5mm,
 xticklabels at=edge bottom,
 yticklabels at=edge left
 },
 scaled x ticks = false,
 x tick label style={
 /pgf/number format/.cd,
 fixed,
 fixed zerofill,
 int detect,
 1000 sep={},
 precision=3
 },
 ylabel style={
 				align=center
 			},
 every axis plot/.append style={thick},
 ymin=0,ymax=550,
 xmin=0,xmax=10000
 ]
 \nextgroupplot[
 legend columns=3,
 legend entries={ME, GE, Threshold},
 legend to name=cp_fitLegend,
 ylabel={Reward},
 xlabel={Number of solutions},
 title={}
 ]
 \linewitherror{data/cp_fit.txt}{b}{mem}{mes}{cola1}
 \linewitherror{data/cp_fit.txt}{b}{gem}{ges}{cola2}
 \addplot [dashed, ultra thick, black] coordinates {(0, 475) (10000, 475)};
 \end{groupplot}
 \end{tikzpicture}
 \pgfplotslegendfromname{cp_fitLegend}
 \caption{Fitness trends on the Cart Pole environment with ME (red) and GE (blue). The dashed line indicates the ``solved'' threshold.}
 \label{fig:cp_fit}
 \end{figure}
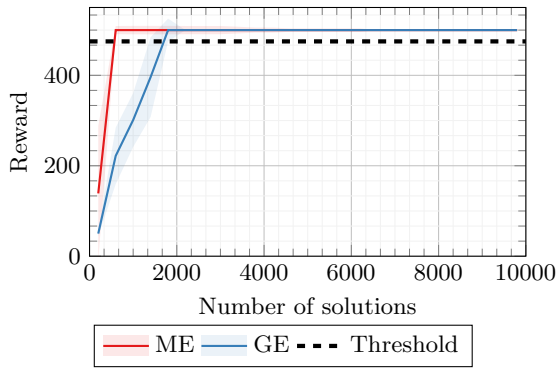

As introduced before, we compare the results from both a performance and a diversity point of view. 
\Cref{fig:cp_fit} shows the trends of the best solutions found during the evolution. Both EAs produce solutions capable to solve the task in less than $2000$ fitness evaluations. 
Of note, ME solves the task faster than GE, in terms of number of fitness evaluations.
 
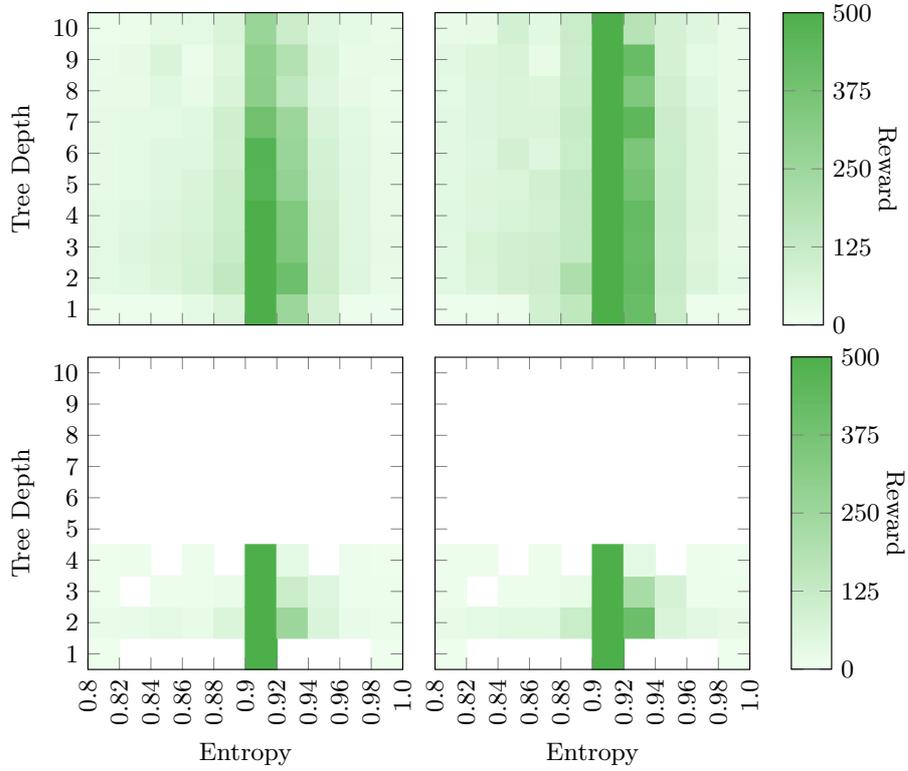
\begin{figure}[ht!]
 \centering
 \resizebox{\textwidth}{!}{
 \begin{tikzpicture}
 \begin{groupplot}[
 width=0.45\linewidth,
 height=0.45\linewidth,
 group style={
 group size=2 by 2,
 horizontal sep=4mm,
 vertical sep=4mm
 },
 title style={anchor=north, yshift=2ex},
 ymin=-0.5,ymax=9.5,
 xmin=-0.5,xmax=9.5,
 axis on top,
 enlargelimits=false,
 point meta=explicit,
 mesh/cols=10,
 mesh/rows=10,
 xticklabels={0.8, 0.82, 0.84, 0.86, 0.88, 0.9, 0.92, 0.94, 0.96, 0.98 ,1.0},
 xtick={-0.5, 0.5, 1.5, 2.5, 3.5, 4.5, 5.5, 6.5, 7.5, 8.5, 9.5},
 x tick label style={rotate=90,anchor=east},
 yticklabels={1,...,10},
 ytick={0,...,9}
 ]
 \nextgroupplot[title={},colormap name=mapCol3, xticklabels={,,}, point meta min=0,point meta max=500, ylabel={Tree Depth}]
 \addplot [matrix plot*] table [meta=a] {data/CartPole-ME_08.txt};
 \nextgroupplot[title={},colormap name=mapCol3 ,point meta min=0,point meta max=500, colorbar right, xticklabels={,,},yticklabels={,,}, colorbar style={ylabel=Reward, ticks=major,every axis y label/.style={at={(0,0.5)}, xshift=4em,rotate=-90}, ytick={0,125,250,375,500}}]
 \addplot [matrix plot*] table [meta=m] {data/CartPole-ME_08.txt};
 
 \nextgroupplot[title={},colormap name=mapCol3,point meta min=0,point meta max=500, ylabel={Tree Depth}, xlabel={Entropy}]
 \addplot [matrix plot*] table [meta=a] {data/CartPole-GE_08.txt};
 \nextgroupplot[title={},colormap name=mapCol3,point meta min=0,point meta max=500, colorbar right, yticklabels={,,}, xlabel={Entropy}, colorbar style={ylabel=Reward, ticks=major,every axis y label/.style={at={(0,0.5)}, xshift=4em,rotate=-90}, ytick={0,125,250,375,500}}]
 \addplot [matrix plot*] table [meta=m] {data/CartPole-GE_08.txt};
 \end{groupplot}
 \end{tikzpicture}
 }
 \caption{Maps obtained with ME (top row) and GE (bottom row) on the Cart Pole environment. In the left column the results in each bin are averaged over $5$ independent runs. Instead, in the right column each bin shows the maximum fitness over $5$ runs.}
 \label{fig:cp-maps}
\end{figure}

\begin{figure}[ht!]
 \newcommand{\x}{0.49}
 \centering
 \begin{subfigure}[b]{\x\textwidth}
 \begin{adjustbox}{width=\textwidth}
 \begin{tikzpicture}[sibling distance=15em]
 \node [shape=rectangle, rounded corners,
 draw, align=center] {$-0.233x_0 -0.753x_1 +$\\$-0.842x_2 -0.919x_3 +< 0.008$}
 child {node [shape=rectangle, rounded corners,
 draw, align=center] {Push\\Left} edge from parent [black] node [left, xshift=-5pt] {False}}
 child {node [shape=rectangle, rounded corners,
 draw, align=center] {Push\\Right} edge from parent [black] node [right, xshift=5pt] {True}};
 \end{tikzpicture}
 \end{adjustbox}
 \caption{Example tree evolved with ME.}
 \end{subfigure}
 \hfill
 \begin{subfigure}[b]{\x\textwidth}
 \begin{adjustbox}{width=\textwidth}
 \begin{tikzpicture}[sibling distance=15em]
 \node [shape=rectangle, rounded corners,
 draw, align=center] {$0.239x_0 +0.171x_1 +$\\$+0.73x_2 +0.36x_3 < -0.059$}
 child {node [shape=rectangle, rounded corners,
 draw, align=center] {Push\\Right} edge from parent [black] node [left, xshift=-5pt] {False}}
 child {node [shape=rectangle, rounded corners,
 draw, align=center] {Push\\Left} edge from parent [black] node [right, xshift=5pt] {True}};
 \end{tikzpicture}
 \end{adjustbox}
 \caption{Example tree evolved with GE.}
 \end{subfigure}
 \caption{Representation of two decision trees that solve the Cart Pole environment (after simplification). 
 Both EAs are able to find solutions that solve the task based on a single condition.}
 \label{fig:treecp}
\end{figure}
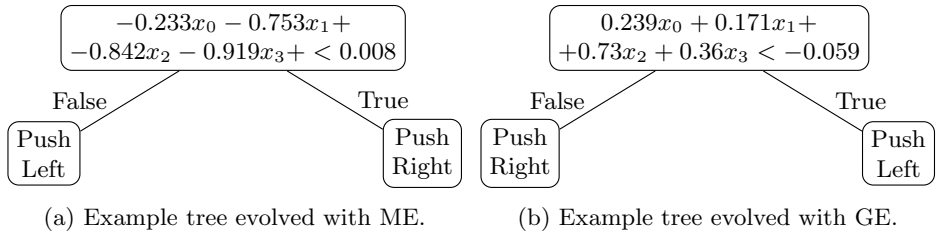

Concerning the illumination capability of the algorithms,~\Cref{fig:cp-maps} shows the archives at the end of the evolution for ME and GE. Note that, in the case of GE, we consider all the individuals generated during the evolutionary process, rather than just the last generation, and fill the map a posteriori. In the case of ME, instead, the map is filled during the evolutionary process, by construction of this algorithm. The results show that, while GE can find solutions that solve the task, as expected its ability to illuminate the feature space is limited, as the algorithm does not allow to find a sufficient number of diverse solutions. On the other hand, ME finds at least one solution for each possible tree depth.~\Cref{fig:treecp} shows two example decision trees that solve the task.
 
Regarding the behavioral feature, while ME still finds more different and high-performing solutions, both EAs seem to focus on entropy values around $0.9-0.92$. 
 
This is probably due to the nature of the environment, which requires high coordination between the two actions (push to the left or the right), leading to a similar frequency for the actions, and, hence, high entropy.
 
\subsection{Mountain Car}
\begin{figure}[ht!]
 \centering
 \begin{tikzpicture}
 \begin{groupplot}[
 width=0.6\linewidth,
 height=0.4\linewidth,
 grid=both,
 grid style={line width=.1pt, draw=gray!10},
 major grid style={line width=.2pt,draw=gray!50},
 minor tick num=5,
 group style={
 group size=1 by 1,
 horizontal sep=4mm,
 vertical sep=5mm,
 xticklabels at=edge bottom,
 yticklabels at=edge left
 },
 scaled x ticks = false,
 x tick label style={
 /pgf/number format/.cd,
 fixed,
 fixed zerofill,
 int detect,
 1000 sep={},
 precision=3
 },
 ylabel style={
 			align=center
 		},
 every axis plot/.append style={thick},
 ymin=-200,ymax=-80,
 xmin=0,xmax=200000
 ]
 \nextgroupplot[
 legend columns=3,
 legend pos=outer north east,
 legend entries={ME, GE, Threshold},
 legend to name=mc_fitLegend,
 ylabel={Reward},
 xlabel={Number of solutions},
 title={}
 ]
 \linewitherror{data/mc_fit.txt}{b}{mem}{mes}{cola1}
 \linewitherror{data/mc_fit.txt}{b}{gem}{ges}{cola2}
 \addplot [dashed, ultra thick, black] coordinates {(0, -110) (200000, -110)};
 \end{groupplot}
 \end{tikzpicture}
 \pgfplotslegendfromname{mc_fitLegend}
 \caption{Fitness trends on the Mountain Car environment with ME (red) and GE (blue). The dashed line indicates the ``solved'' threshold.}
 \label{fig:mc_fit}
 \end{figure}
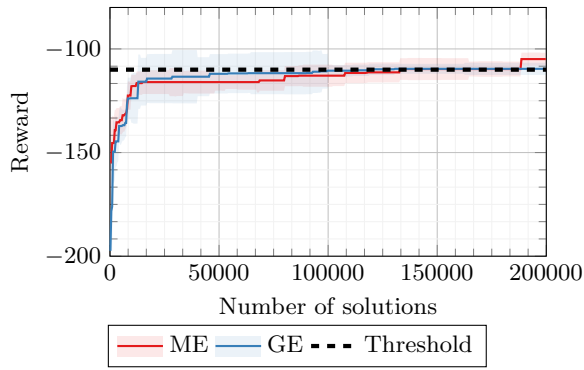

As regards the Mountain Car task,~\Cref{fig:mc_fit} shows the fitness trend for the two algorithms. 
 
As in the previous case, both algorithms can solve the task. However, in this case, GE is faster at solving the task than ME: the former needs around $110000$ fitness evaluations; the latter, instead, finds the first solution that solves the task after around $130000$ fitness evaluations.
 
In other words, while eventually reaching slightly better performance, ME requires $10\%$ of the total amount of fitness evaluations budget more than GE to solve the task. 
 
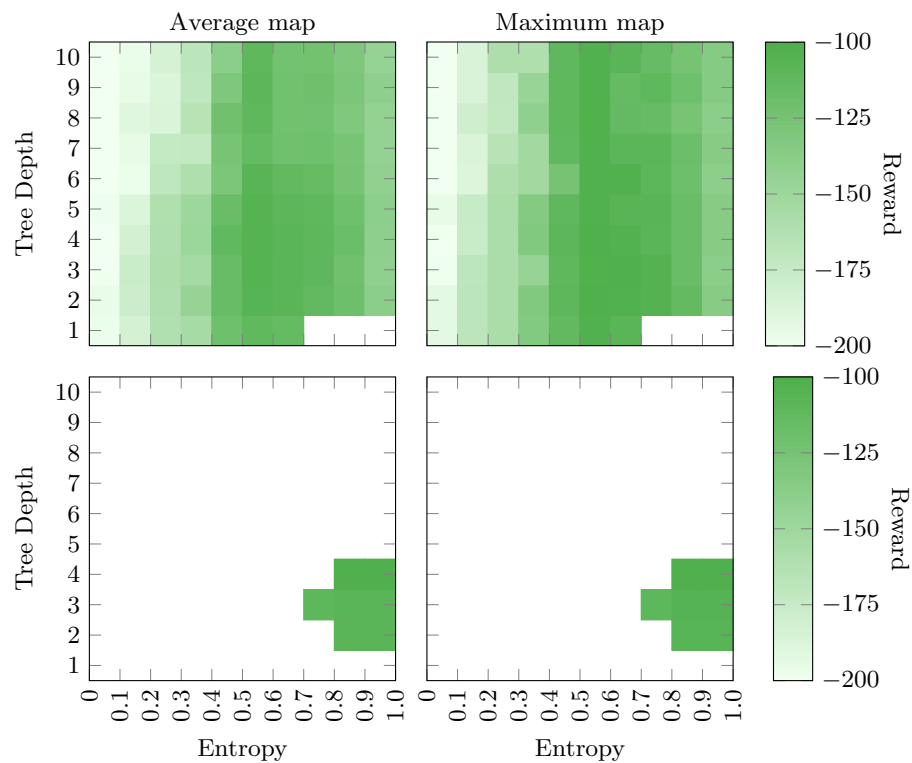
\begin{figure}[ht!]
 \centering
 \resizebox{\textwidth}{!}{
 \begin{tikzpicture}
 \begin{groupplot}[
 width=0.45\linewidth,
 height=0.45\linewidth,
 group style={
 group size=2 by 2,
 horizontal sep=4mm,
 vertical sep=4mm
 },
 title style={anchor=north, yshift=2ex},
 ymin=-0.5,ymax=9.5,
 xmin=-0.5,xmax=9.5,
 axis on top,
 enlargelimits=false,
 point meta=explicit,
 mesh/cols=10,
 mesh/rows=10,
 xticklabels={0, 0.1, 0.2, 0.3, 0.4, 0.5, 0.6,0.7,0.8,0.9,1.0},
 xtick={-0.5, 0.5, 1.5, 2.5, 3.5, 4.5, 5.5,6.5,7.5,8.5,9.5},
 x tick label style={rotate=90,anchor=east},
 yticklabels={1,...,10},
 ytick={0,...,9}
 ]
 \nextgroupplot[title={Average map},colormap name=mapCol3, xticklabels={,,}, point meta min=-200,point meta max=-100, ylabel={Tree Depth}]
 \addplot [matrix plot*] table [meta=a] {data/MountainCar-ME.txt};
 \nextgroupplot[title={Maximum map},colormap name=mapCol3 ,point meta min=-200,point meta max=-100, colorbar right, xticklabels={,,},yticklabels={,,}, colorbar style={ylabel=Reward, ticks=major,every axis y label/.style={at={(0,0.5)}, xshift=5em,rotate=-90}, ytick={-200, -175, -150, -125, -100}}]
 \addplot [matrix plot*] table [meta=m] {data/MountainCar-ME.txt};
 
 \nextgroupplot[title={},colormap name=mapCol3,point meta min=-200,point meta max=-100, ylabel={Tree Depth}, xlabel={Entropy}]
 \addplot [matrix plot*] table [meta=a] {data/MountainCar-GE.txt};
 \nextgroupplot[title={},colormap name=mapCol3,point meta min=-200,point meta max=-100, colorbar right, yticklabels={,,}, xlabel={Entropy}, colorbar style={ylabel=Reward, ticks=major,every axis y label/.style={at={(0,0.5)}, xshift=5em,rotate=-90}, ytick={-200, -175, -150, -125, -100}}]
 \addplot [matrix plot*] table [meta=m] {data/MountainCar-GE.txt};
 \end{groupplot}
 \end{tikzpicture}
 }
 \caption{Maps obtained with ME (top row) and GE (bottom row) on the Mountain Car environment. In the left column the results in each bin are averaged over $5$ independent runs. Instead, in the right column each bin shows the maximum fitness over $5$ runs.}
 \label{fig:mc-maps}
\end{figure}

\Cref{fig:mc-maps} shows the archive at the end of the evolution for the two EAs.
Similar to the Cart Pole case, ME illuminates the feature space better than GE, covering $97\%$ of bins in all $5$ runs. On the other hand, GE concentrates on a small portion of the feature space. 
Overall, we can observe that the two algorithms find solutions that solve the problem in different areas of the feature space.
Regarding the behavioral feature, while GE trees present a high entropy level as in the Cart Pole task, ME produces also trees that have lower entropy. Hence, these trees present behaviors in which at least one action is less frequent than the others.
For the structural feature, we can observe that, as in the Cart Pole environment, GE focus on small trees, while ME produces solutions that cover the entire range $[1,10]$.
 
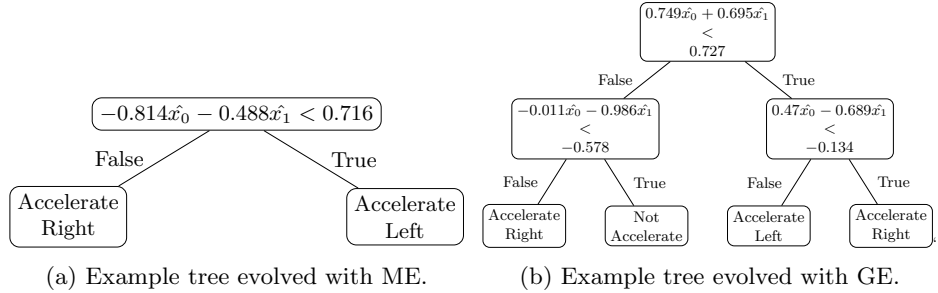
\begin{figure}[ht!]
 \newcommand{\x}{0.49}
 \centering
 \begin{subfigure}[b]{\x\textwidth}
 \begin{adjustbox}{width=\textwidth}
 \begin{tikzpicture}[sibling distance=15em]
 \node [shape=rectangle, rounded corners,
 draw, align=center] {$-0.814\hat{x_0} -0.488\hat{x_1} < 0.716$}
 child {node [shape=rectangle, rounded corners,
 draw, align=center] {Accelerate\\Right} edge from parent [black] node [left, xshift=-5pt] {False}}
 child {node [shape=rectangle, rounded corners,
 draw, align=center] {Accelerate\\Left} edge from parent [black] node [right, xshift=5pt] {True}};
 \end{tikzpicture}
 \end{adjustbox}
 \caption{Example tree evolved with ME.}
 \end{subfigure}
 \hfill
 \begin{subfigure}[b]{\x\textwidth}
 \begin{adjustbox}{width=\textwidth}
 \begin{tikzpicture}[auto,
 level 1/.style={sibling distance=5cm, level distance=2cm},
 level 2/.style={sibling distance=2.5cm, level distance=2cm}]
 \node [shape=rectangle, rounded corners, draw, align=center] {$0.749\hat{x_0}+0.695\hat{x_1}$\\$<$\\ $0.727$}
 child {node [shape=rectangle, rounded corners, draw, align=center] {$-0.011\hat{x_0} -0.986\hat{x_1}$ \\$<$ \\$-0.578$}
 child{ node [shape=rectangle, rounded corners, draw, align=center] {Accelerate\\Right} edge from parent [black] node [left, xshift=-5pt] {False}}
 child{ node [shape=rectangle, rounded corners, draw, align=center] {Not\\Accelerate} edge from parent [black] node [right, xshift=5pt] {True}}
 edge from parent [black] node [left, xshift=-5pt] {False}}
 child {node [shape=rectangle, rounded corners, draw, align=center] {$0.47\hat{x_0}-0.689\hat{x_1}$\\$<$\\$-0.134$} 
 child{ node [shape=rectangle, rounded corners, draw, align=center] {Accelerate\\Left} edge from parent [black] node [left, xshift=-5pt] {False}}
 child{ node [shape=rectangle, rounded corners, draw, align=center] {Accelerate\\Right} edge from parent [black] node [right, xshift=5pt] {True}}
 edge from parent [black] node [right, xshift=5pt] {True}};
 \end{tikzpicture}`
 \end{adjustbox}
 \caption{Example tree evolved with GE.}
 \end{subfigure}
 \caption{Representation of two decision trees that solve the Mountain Car environment (after simplification). GE (right) finds solutions that use all the tree actions (see~\Cref{subsec:mc}). Hence, the depth of the tree is $2$. Meanwhile, ME (left) finds also solutions that do not use the \emph{not accelerate} action. Therefore it is possible to produce a decision tree with a depth of $1$.}
 \label{fig:treemc}
\end{figure}

In particular, ME produces also trees with a depth equal to $1$, meaning that the maximum number of leaves is $2$. Hence, the entropy in this case is limited to a maximum of circa $0.63$, corresponding to the case in which the two actions are executed an equal number of times (we remember that we calculate the entropy using as the base for the logarithm the number of actions, see~\Cref{subsec:descriptor}).~\Cref{fig:treemc} shows a representation of two example trees.
 
\section{Conclusion}
\label{sec:conclusions}

In this paper, we have applied a QD algorithm, namely ME, for finding a diverse collection of interpretable hybrid models composed of a decision tree combined with Q-learning on the leaves. We have tested the method on two tasks from OpenAI Gym library, namely Cart Pole and Mountain Car, and compared the results of ME with those obtained by GE. We have then discussed the results of the two algorithms in terms of performance and ``illumination'' capability, given a feature space defined by model complexity and behavioral variability.

Summarizing, we observed that, in both tasks, ME finds solutions that solve the task, ``illuminating'' at the same time the feature space in a more efficient way w.r.t. GE. Moreover, while both EAs produced models with low complexity, hence good interpretability, in the Mountain Car environment ME found that one action is not necessary to solve the task.

In future works, we will extend this study to more recent variants of ME, such as those proposed in \cite{vassiliades2017using,fontaine2020covariance}, and to more challenging RL tasks. Moreover, we will investigate the scalability of ME w.r.t. the number of features used in the descriptor. Another interesting direction would be to introduce interactions with the user during the search process, as done in \cite{urquhart2019increasing}.


\bibliographystyle{splncs} 
\bibliography{main,main-xai}



\end{document}